\documentclass[runningheads]{llncs}
\usepackage[T1]{fontenc}
\usepackage{enumerate,multirow,subcaption,hyperref,enumitem, amssymb, amsmath, booktabs}
\usepackage[misc]{ifsym} 
\usepackage{graphicx}
\usepackage{color}

\begin{document}
\title{STMGF: An Effective Spatial-Temporal Multi-Granularity Framework for Traffic Forecasting}
\titlerunning{STMGF}
%
%
\author{Zhengyang Zhao\inst{1,2}\and
Haitao Yuan\inst{3}$^{(\textrm{\Letter})}$ \and
Nan Jiang\inst{1}\and
Minxiao	Chen\inst{1}\and
Ning Liu\inst{2}$^{(\textrm{\Letter})}$\and
Zengxiang Li\inst{4}
}
\authorrunning{Z. Zhao et al.}
%
\institute{Beijing University of Posts and Telecommunications, Beijing, China\and
School of Software, Shandong University, Shandong, China\and
Nanyang Technological University, Singapore\and
Digital Research Institute
of ENN Group, Beijing, China\\
\email{\{zhaozhengyang,chenminxiao\}@bupt.edu.cn, yht19@tsinghua.org.cn, jn\_work@outlook.com, liun21cs@sdu.edu.cn, lizengxiang@enn.cn}\\
}
\maketitle              
\begin{abstract}

Accurate Traffic Prediction is a challenging task in intelligent transportation due to the spatial-temporal aspects of road networks. The traffic of a road network can be affected by long-distance or long-term dependencies where existing methods fall short in modeling them. In this paper, we introduce a novel framework known as \textbf{S}patial-\textbf{T}emporal \textbf{M}ulti-\textbf{G}ranularity \textbf{F}ramework (STMGF) to enhance the capture of long-distance and long-term information of the road networks. STMGF makes full use of different granularity information of road networks and models the long-distance and long-term information by gathering information in a hierarchical interactive way. Further, it leverages the inherent periodicity in traffic sequences to refine prediction results by matching with recent traffic data. We conduct experiments on two real-world datasets, and the results demonstrate that STMGF outperforms all baseline models and achieves state-of-the-art performance.

\end{abstract}


\section{INTRODUCTION}
With the development of artificial intelligence technology, traffic forecasting has gradually become a reality, which is significant for people's daily travel, vehicle path planning, and smart city construction. Many researches have been conducted to improve the prediction accuracy. Traditional machine learning methods like VAR~\cite{lippi2013short}, SVR~\cite{smola2004tutorial}, ARIMA~\cite{zhang2003time}, etc., are not excellent enough to handle difficult unexpected situations in real road networks. With the development of deep learning technology, recurrent neural network~\cite{medsker2001recurrent,grossberg2013recurrent} based models such as Gated Recurrent Unit (GRU)~\cite{cho2014properties} and Long Short Term Memory (LSTM)~\cite{hochreiter1997long} are used for multivariate time series prediction. These methods do not take the spatial dependency of traffic sequences into account. Recently, researchers have proposed spatial-temporal graph neural networks (STGNN)~\cite{sahili2023spatio,wang2020traffic} based on graph neural network and time series models, such as DCRNN~\cite{DCRNN}, STGCN~\cite{yu2017spatio}, DSTAGNN~\cite{lan2022dstagnn}. These methods simultaneously model the temporal and spatial features of traffic flow and have achieved encouraging results. 

\begin{figure}[!t]
  \centering
  \includegraphics[width=.7\linewidth]{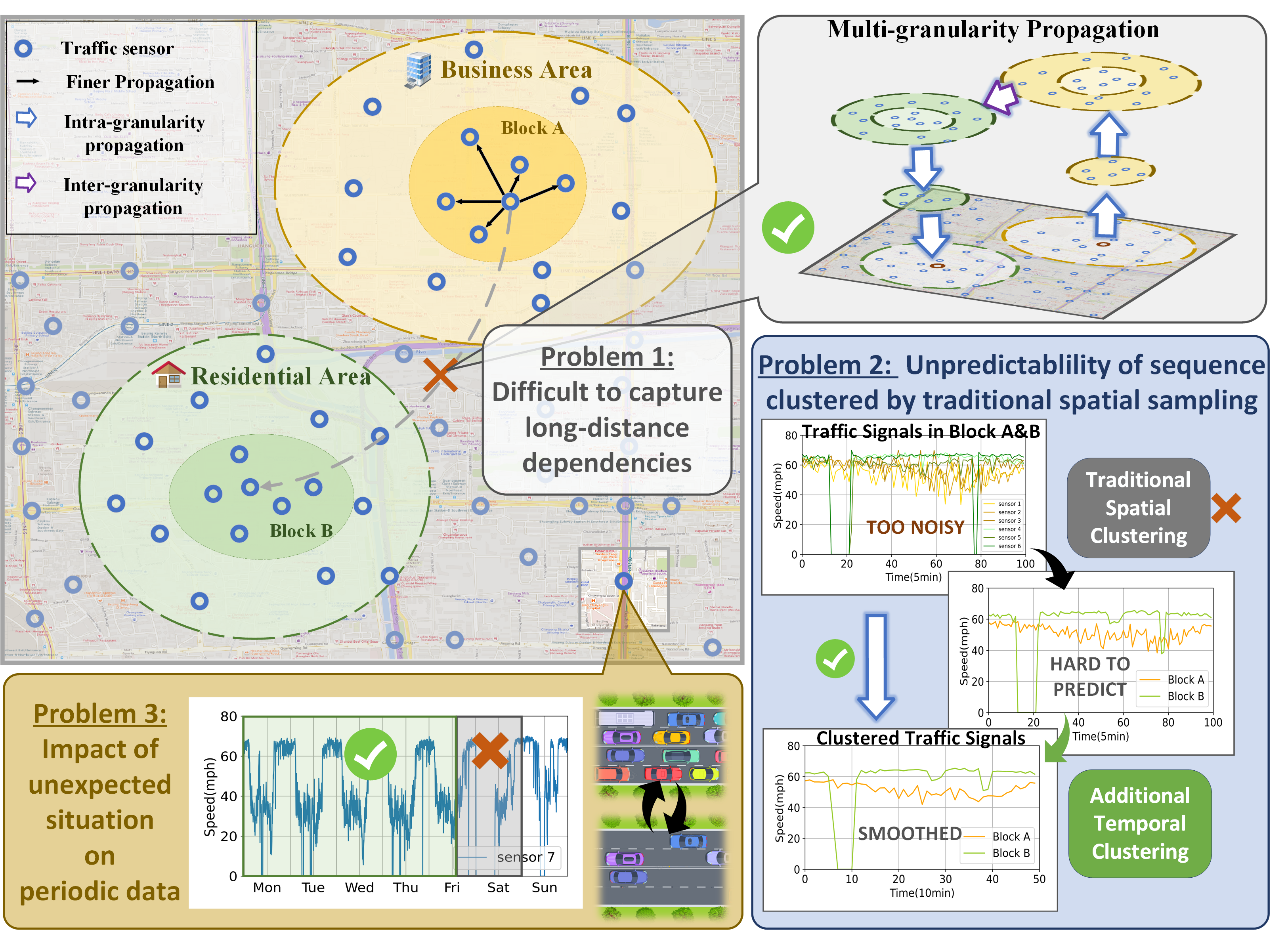}
  \vspace{-0.1in}
  \caption{Three main difficult problems for traffic forecasting.}
  \vspace{-0.3in}
  \label{fig:intro}
\end{figure}

However, as shown in \autoref{fig:intro}, accurate traffic prediction still faces various problems, therefore, we have designed a novel Spatial-Temporal Multi-Granularity Framework (STMGF) to address these issues:

\noindent \textbf{(1) Difficulty in capturing long-distance dependencies:} As distance increases, vehicle interference increases, making it difficult to directly capture the dependency relationships between sensors.
We design a spatial multi-granularity mechanism to enhance the capturing of long-distance dependencies by categorizing the traffic network into sensors, blocks, and urban functional areas. And we exploit inter-granularity and intra-granularity propagation processes to model traffic signal dynamics efficiently. We construct dependencies between distant nodes in a cross-layer manner, reducing propagation hops and enabling better capture of long-distance dependencies among sensors.

\noindent \textbf{(2) Unpredictability of sequence clustered by traditional spatial sampling:} A large number of external signals introduce noise into traffic data. Due to the randomness of noise, spatial clustering of nodes can lead to smooth traffic sequences. A smooth sequence means it is suitable for time sampling with coarser granularity. At this point, if fine-grained time scale sampling is still used, it will cause a bias between the clustered sequence and the original sequence. Therefore, we propose a time aggregation based temporal multi-granularity method which synchronizes with spatial multi-granularity. This approach not only models the traffic sequence more reasonably, but also reduces the number of farthest prediction hops, which enables effective capture of long-term dependencies.

\noindent  \textbf{(3) Ineffectiveness in Utilizing Historical Period Data:} In addition to recent time aggregation, at further time scales, the periodicity of traffic data makes historical period data valuable for predicting results. However, unpredictable sudden events (e.g., abrupt speed drops on a Friday night) can disrupt regular periodicity. Therefore, we design a matching mechanism between historical data and recent data to selectively extract historical patterns. Through similarity matching, we reduce the impact of unpredictable events that disrupt the cycle and optimize the information extraction of historical period data.

In summary, the main contributions of this paper are as follows:
\begin{itemize}
\item We reveal the reasons for the low prediction accuracy of existing methods in long-distance and long-term traffic forecasting and propose three existing problems.
\item We introduce a novel multi-granularity framework to enhance the capture of long-
distance and long-term information of road network.
\item We validate the effectiveness of our proposed framework using two distinct types of real-world datasets and demonstrate that our framework consistently outperforms all baseline models.
\end{itemize}

\vspace{-0.2in}
\begin{figure*}
  \centering
  \includegraphics[width=\linewidth,height=0.4\textwidth]{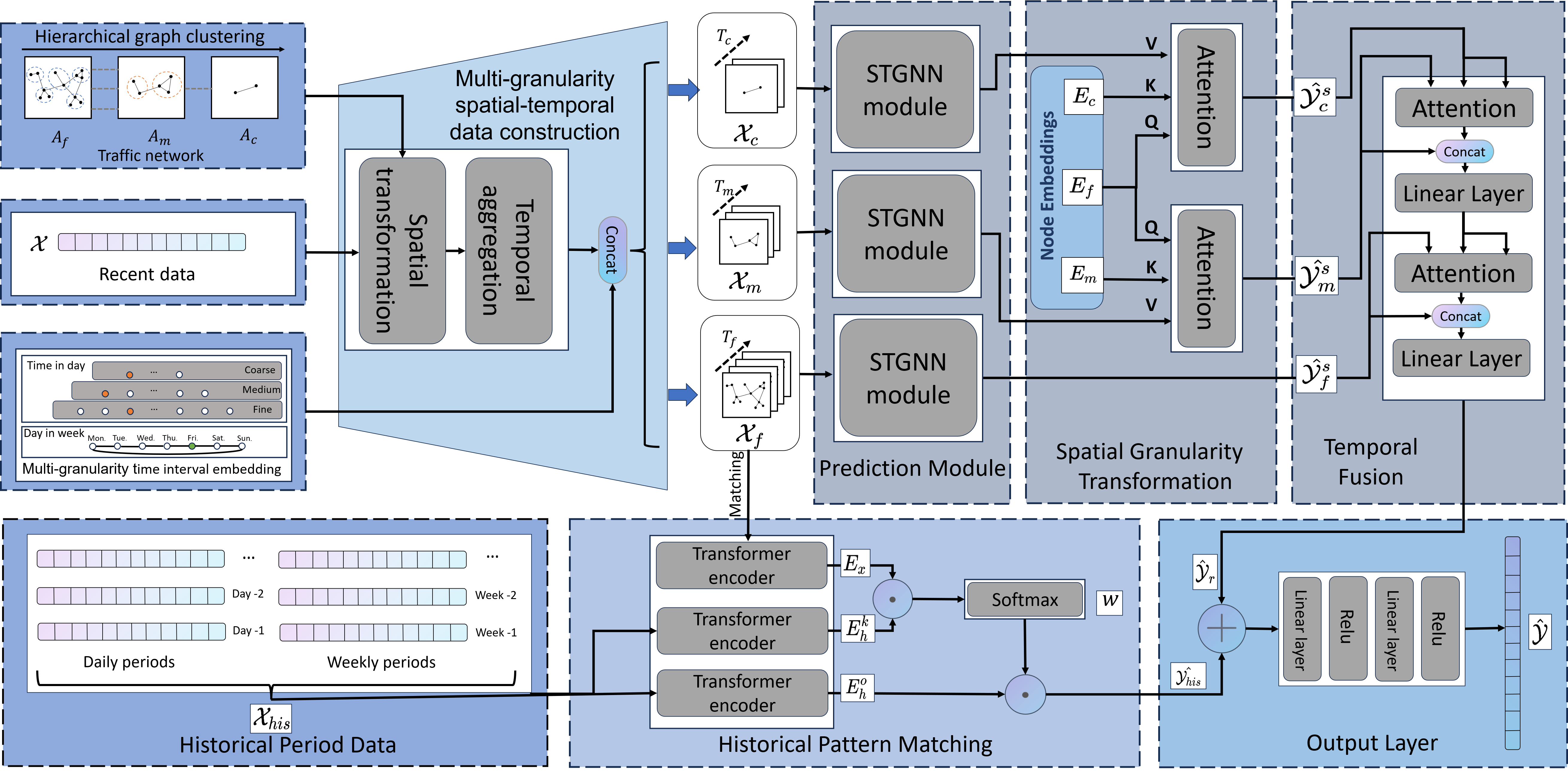}
  \vspace{-0.2in}
  \caption{The overall architecture of the proposed STMGF.}
  \vspace{-0.2in}
  \label{fig:architecture}
\end{figure*}
\vspace{-0.2in}
\section{Problem Formulation}\label{chaper2}

A traffic network \( G = (V, E) \) contains $N=|V|$ traffic sensors positioned on the roadside and their connectivity relationships $E$. Given a traffic network $G$ and a traffic signal tensor \( \mathcal{X}=[X_{t-T+1},\cdots,X_{t}] \in \mathbb{R}^{T \times N \times C} \) collected by the sensors, which represents the data from all $N$ traffic sensors on the network $G$ over the preceding \( T \) time slices before the current time slice $t$. These traffic signals have $|C|$ channels and may encompass variables such as speed, flow, and occupancy. The objective is to forecast the future traffic signal tensor \( \mathcal{Y} = [Y_{t+1}, Y_{t+2}, \ldots, Y_{t+T'}] \in \mathbb{R}^{T' \times N \times C} \) to minimize the difference between $Y_{j}$ and the ground truth $X_j$ for each subsequent time slice $j \in [t+1, t+T]$.

\section{METHODOLOGY}\label{chaper3}
In this section, we first introduce the overall architecture of our framework and then describe the specific design of the proposed components respectively.
\subsection{MODEL OVERVIEW}

As shown in \autoref{fig:architecture}, the input of STMGF is divided into three parts: the adjacency matrix $A$ of the traffic network, recent data $\mathcal{X}$, and historical period data $\mathcal{X}_{his}$. We use the adjacency matrix and recent data to generate $\mathcal{\hat Y}_r$ through multi-granularity prediction process(\autoref{3.2}), which consists of \textit{multi-granularity spatial-temporal data construction}, \textit{prediction module}, \textit{spatial granularity transformation} and \textit{temporal fusion}. We use the historical period data to generate $\mathcal{\hat Y}_{his}$ by \textit{historical pattern matching}. We finally get the output of STMGF in the \textit{output layer}.

\subsection{MULTI-GRANULARITY PREDICTION}\label{3.2}

\noindent\textbf{Multi-granularity spatial-temporal data construction.} 
The construction of multi-granularity spatial-temporal data includes four aspects. For spatial dimensions, we conduct graph hierarchical clustering and spatial transformation. For temporal dimensions, we design temporal aggregation and multi-granularity time interval embedding.

\textbf{(i) Hierarchical clustering.} 
Based on the assumption that each sensor or has the highest degree of interaction with the nearest sensors, we use the distance information acquired from adjacency matrix $A$ to construct the hierarchical graph by the Paris algorithm~\cite{bonald2018hierarchical}, which is based on node pair sampling to obtain hierarchical clustering. 
The original adjacency matrix is converted into three new matrices with coarse, medium, and fine granularity through hierarchical clustering to extract a variety of different information, which are denoted as $A_c$, $A_m$, and $A_f$. We introduce a hyper-parameter clustering factor $f_s$, which controls the number of nodes ($N_c$, $N_m$ and $N_f$) in the three layers to be $f_s$, $f_s^2$, and $N$.

\textbf{(ii) Spatial transformation.} For recent traffic signal $\mathcal{X}_r\in \mathbb{R}^{T_r \times N\times C}$, we design a transformation matrix $Tran$:
\begin{equation}
    \forall i\in N',j\in N\quad Tran_{i,j}=
\begin{cases}
\frac{1}{|i|}& \text{j $\in$  i}\\
0& \text{otherwise}
\end{cases}
\end{equation}

\noindent where $N'$ denotes the set of clusters on the clustered graph. The transformation matrix uses the method of taking the average within the cluster, which is beneficial for capturing overall information and reducing random errors. Hence, we can generate the traffic signal of spatial multi-granularity from $\mathcal{X}$:
\begin{equation}
\begin{aligned}
    \mathcal{X'}_c &= Tran_c \mathcal{X}\\
    \mathcal{X'}_m &= Tran_m \mathcal{X}
\end{aligned}
\end{equation}

\noindent where $\mathcal{X'}_c \in \mathbb{R}^{T_r \times N_c\times C}$,  $\mathcal{X'}_m \in \mathbb{R}^{T_r \times N_m\times C}$($'$ denotes traffic signals that have not undergone temporal aggregation). Besides, we simply use $\mathcal{X}_f=\mathcal{X}$ for the fine-grained prediction. 

\begin{figure*}[htbp]
	\centering
	\subfloat[Temporal aggregation.]{\includegraphics[width=.39\linewidth]{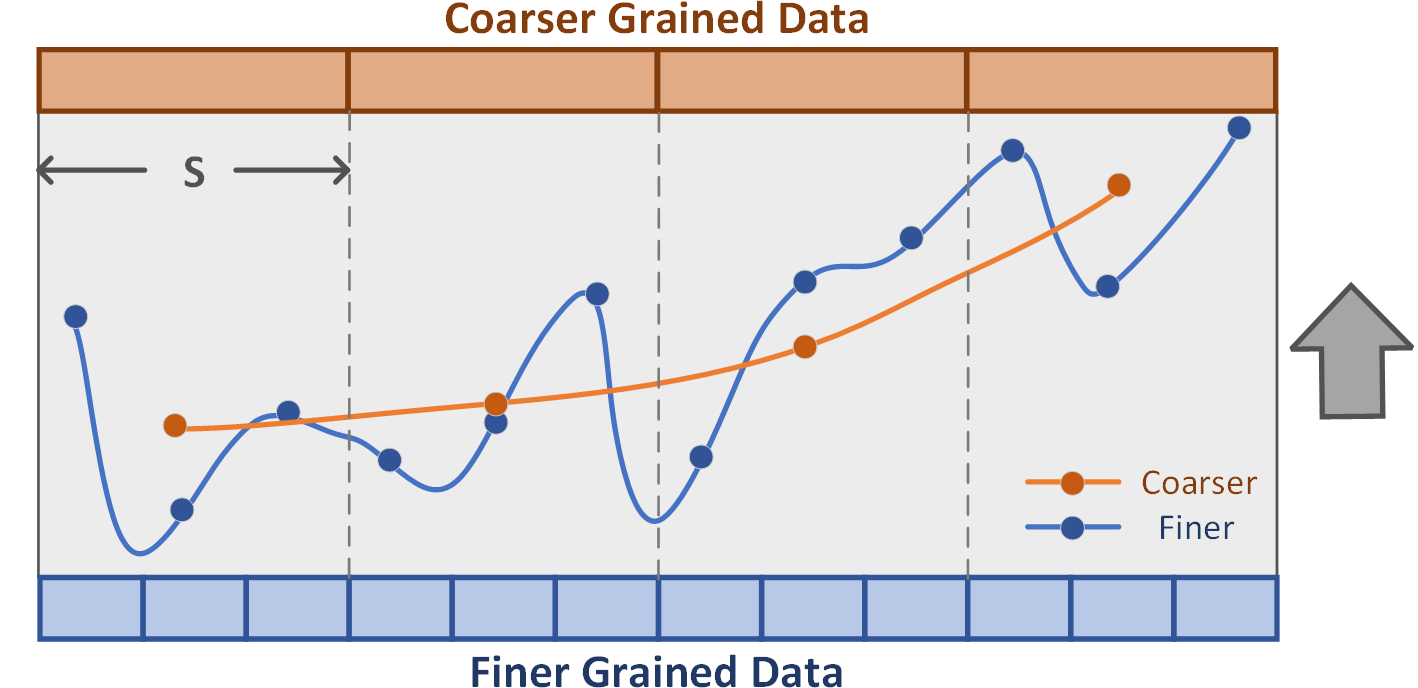}\label{fig:temporal}}\hspace{10pt}
	\subfloat[Spatial granularity transformation.]{\includegraphics[width=.54\linewidth]{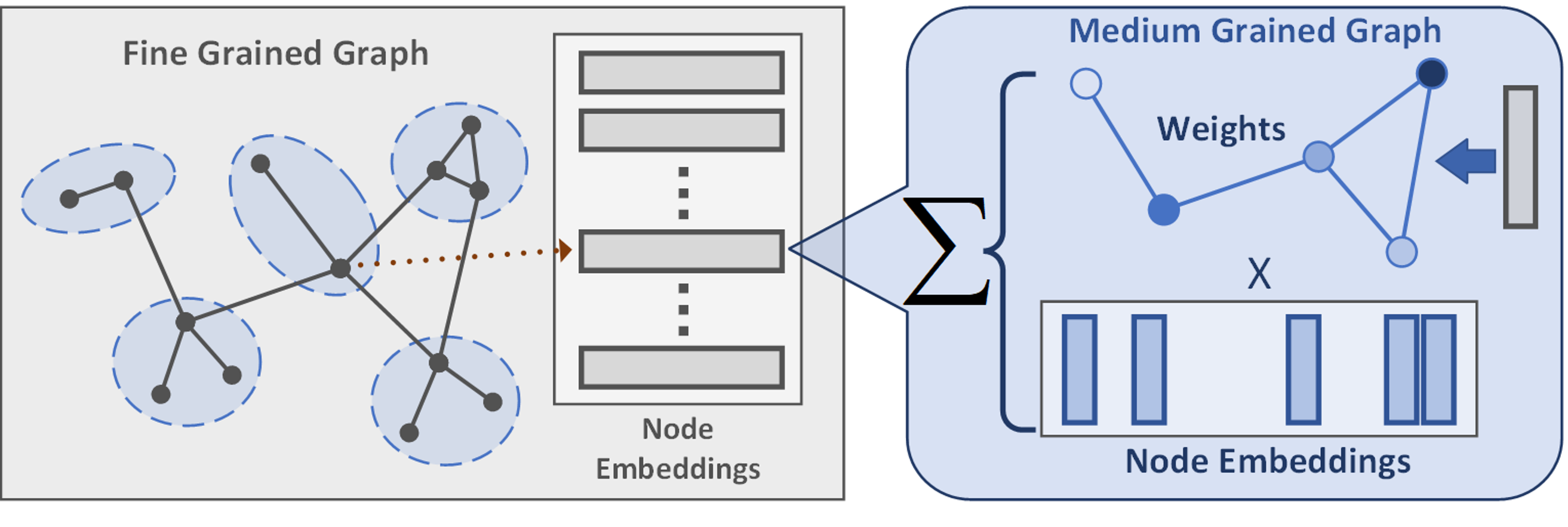}\label{fig:spatial}}
    \caption{Temporal aggregation and spatial granularity transformation}
  \vspace{-0.3in}
\end{figure*}

\textbf{(iii) Temporal aggregation.} 
As shown in \autoref{fig:temporal}, we divide the recent data into several segments $S$ and generate a new representation $S_1, \dots, S_i$ of recent data from the segment level. Due to the continuity of the time series, we design a piecewise averaging method to generate the coarser temporal granularity data $\mathcal{X}_c\in \mathbb{R}^{T_c\times N_c\times C}$ and $\mathcal{X}_m\in \mathbb{R}^{T_m\times N_m\times C}$  from $\mathcal{X'}_c$, $\mathcal{X'}_m$:
\begin{equation}
\begin{aligned}
    \mathcal{X}_c&=||_{i=1}^{T_c}\text{mean}\{S^c_i\}\\
    \mathcal{X}_m&=||_{i=1}^{T_m}\text{mean}\{S^m_i\}
\end{aligned}
\end{equation}
\noindent where $||$ means the concatenation operation and mean denotes the average operation. To control the granularity, we introduce hyper-parameters $T_c$ and $T_m$ to determine the number of aggregated time slices. In this way, the number of farthest prediction hops has changed from the previous $T_f+T_F-1$ to $T_c+T_c-1$ and $T_m+T_m-1$ steps, greatly improving the prediction accuracy of the two coarser-grained layers. 

\textbf{(iv) Multi-granularity time interval embedding.} 
We design two types of time interval embeddings of different granularities as features to enhance the acquisition of temporal periodicity caused by human activities: (a) day-in-week embedding; (b) time-in-day embedding. The first embedding method uses 1 to 7 to represent different days within a week (Monday to Sunday). The second type divides a day into $\frac{24\times 60}{\Delta t}$ time intervals ($\Delta t$ is a fixed sampling interval such as 5 min). Each time interval corresponds to an embedded number ($1$ to $\frac{24\times 60}{\Delta t}$).

\noindent\textbf{Prediction module.}
In the prediction module, we use the STGNN-based models for multi-granularity prediction. 
We will test the performance of different prediction modules in \autoref{ablation}. For the constructed multi-granularity data $\mathcal{X}_c, \mathcal{X}_m, \mathcal{X}_f$, we feed them into three prediction modules. Then we obtain the results $\mathcal{\hat Y'}_c, \mathcal{\hat Y'}_m, \mathcal{\hat Y'}_f$ of different granularity.

\noindent\textbf{Spatial granularity transformation.} 
The prediction results of coarser granularity can capture long-range node relationships and have lower error levels, which can be used to correct the prediction results of finer granularity. To this end, we design the transformation process (shown in \autoref{fig:spatial}). We use node embedding vectors to extract spatial features and as key and query vectors to construct an attention matrix\cite{vaswani2017attention}. Specifically, for the three-layer graphs with different granularity, we can learn a latent positional embedding vector $E_c \in \mathbb{R}^{N_c \times d_s}$, $E_m \in \mathbb{R}^{N_m \times d_s}$, $E_f \in \mathbb{R}^{N_f \times d_s}$ to model the positional features of each node, where $d_s$ is the dimension of hidden features. We can then construct the attention matrix $M_c \in \mathbb{R}^{N_f \times N_c}$(coarse to fine), $M_m \in \mathbb{R}^{N_f \times N_m}$(middle to fine) from these vectors:
\begin{equation}\label{4}
\begin{aligned}
    M_c&=softmax(\sigma(E_fE_c))\\
    M_m&=softmax(\sigma(E_fE_m))
\end{aligned}
\end{equation}

Given the attention matrix $M_c$ and $M_m$, we can obtain $\mathcal{\hat Y}_c^s=M_c\mathcal{\hat Y'}_c$ and $\mathcal{\hat Y}_m^s=M_m\mathcal{\hat Y'}_m$ as the final prediction result through spatial multi-granularity mechanism .

\noindent\textbf{Temporal fusion.}The coarser granularity data can reflect the changing trend of the time series more accurately. This feature can be utilized to constrain fine-grained prediction results. We design a top-down cross-attention mechanism to fuse time series with different granularities. Specifically, for the predicted results $\mathcal{\hat Y}^s_c$, and $\mathcal{\hat Y}^s_m$,  given the learnable project matrices $W^Q, W^K, W^V \in \mathbb{R}^{d_s\times d_s}$ ($d_s$ represents the hidden features), the cross attention function can be written as:
\begin{equation}
\begin{aligned}
    attn(\mathcal{\hat Y}^s_c, \mathcal{\hat Y}^s_m)&=score\times \mathcal{\hat Y}^s_cW^V\\
    &=softmax(\frac{\mathcal{\hat Y}^s_mW^Q(\mathcal{\hat Y}^s_cW^K)^T}{\sqrt{d}})\mathcal{\hat Y}^s_cW^V
\end{aligned}    
\end{equation}
After the cross-attention processing, the information is concatenated with the lower granularity level data $\mathcal{\hat{Y}}^s_m$ and processed through a linear layer:
\begin{equation}
\begin{aligned}
    \mathcal{\hat Y}_m=\sigma((\mathcal{\hat Y}^s_m||attn(\mathcal{\hat Y}^s_c, \mathcal{\hat Y}^s_m))W)
\end{aligned}    
\end{equation}
\noindent where $W\in \mathbb{R}^{2d_s\times d_s}$. The fusion process between $\mathcal{\hat Y}_m$ and $\mathcal{\hat Y'}_f$ follows the same function.

\subsection{HISTORICAL PATTERN MATCHING}
We propose three steps to process historical period sequences: (1) Encoding the data sequence (2) Querying the correlation between recent data and historical period data. (3) Weighting the historical period sequence with weight vector $w$ to obtain the final representation.
We use transformer encoders\cite{vaswani2017attention} to encode traffic sequences $\mathcal{X}$ and $\mathcal{X}_{his}[i]$ to get $E_x$ and $E_h^k[i]$. Then we compute the matching weight vector $w$ between vector $E_x$ and each encoded vector $E_h^k[i]$ by taking the inner product followed by a softmax operation:
\begin{equation}
    w_i=softmax(E_x^TE_h^k[i])
\end{equation}
The weight vector $w$ is used for weighting historical period data patterns. Then we use another transformer encoder to get the output representation $E_h^o$ of $\mathcal{X}_{his}$ and get $\mathcal{\hat Y}_{his}$ as follows:
\begin{equation}\label{12}
    \mathcal{\hat Y}_{his}=\sum_i w[i]\times E_h^o[i]
\end{equation}

Subsequently, we take $\mathcal{\hat Y}_{r}+\mathcal{\hat Y}_{his}$ as the result and feed them into an output layer. The output layer we designed is a two-layer fully connected layer, using ReLU as the activation function. Therefore, we can get the final result $\mathcal{\hat Y}$ of our STMGF as follows:
\begin{equation}
\mathcal{\hat Y}=\sigma((\sigma((\mathcal{\hat Y}_{r}+\mathcal{\hat Y}_{his})W_1))W_2)
\end{equation}
\noindent where $W_1\in \mathbb{R}^{d\times od}$, $W_2 \in \mathbb{R}^{od\times C_{out}}$ and $od$ denotes the output layer hidden dimension.

\section{EXPERIMENTS}\label{chapter5}
In this section, we demonstrate the experiment results on the two datasets. We first give the experimental settings, and then show the main experiment results compared with baseline methods. Next, we conduct a parameter sensitivity study and ablation study to further validate the properties of different components in STMGF. We further conduct a case study to demonstrate the result.

\subsection{Experimental settings}\label{chapter4}
\textbf{Datasets} We conduct experiments on two commonly used real-world datasets: (1) METR-LA, 4 months of speed data from the road network of Los Angeles County\cite{jagadish2014big}, which has 207 sensors and 34272 time slices; (2) PEMS08, 2 months of flow data from CalTrans PeMS\cite{chen2001freeway}, which has 170 sensors and 17833 time slices.
For a fair comparison, we use the data partitioning method from previous works. For the METR-LA dataset, we use 70\% data for training, 20\% for testing and the remaining 10\% for validation~\cite{DCRNN,wu2019graph}. For the PEMS08 dataset, we use 60\% data for training, 20\% data for testing and the remaining 20\% for validation~\cite{guo2021learning,guo2019attention,song2020spatial}. For the time series data, we use a sliding window with a length of 48 for data partitioning. For each window data, the first 24 time slices are used for input and the last 24 time slices are used for prediction.

\textbf{Baselines}
We choose abundant baseline methods, as well as the very recent state-of-the-art works:(1) Traditional statistical models for traffic forecasting (HA, ARIMA\cite{zhang2003time}) (2) Deep learning models (DCRNN\cite{DCRNN}, GraphWaveNet\cite{wu2019graph}, DSTAGNN\cite{lan2022dstagnn}, ASTGCN\cite{guo2019attention}, PGCN\cite{shin2022pgcn}, HGCN\cite{guo2021hierarchical}, D2STGNN\cite{shao2022decoupled}, MTGNN\cite{wu2020connecting}, PDFormer\cite{jiang2023pdformer}). 

\textbf{Metrics}
To evaluate the predictive accuracy of various methods, we utilize MAE (Mean Absolute Error), RMSE (Root Mean Squared Error), and MAPE (Mean Absolute Percentage Error) as our chosen metrics.

\textbf{Model Settings}
We consider the following hyper-parameters: (1) the clustering factor $f_s$ (2) the length of traffic sequences of different granularity $T_f$, $T_m$ and $T_c$ (3) the number of historical period data $n_d$, $n_w$ (4) the hidden feature $d_s$, $d_t$ and $d_h$ (5) the hidden dimension of the output layer $od$.  For the METR-LA dataset we set $f_s=10, T_f=24, T_m=8, T_c=4, n_d=0, n_w=2, d_s=256, d_t=256, d_h=64, od=512$. 
For the PEMS08 dataset we set $f_s=7, T_f=24, T_m=6, T_c=2, n_d=1, n_w=1, d_s=256, d_t=256, d_h=64, od=512$.
All the experiments of our proposed model are conducted by Pytorch 2.0.1 on a Tesla V100 GPU. We use the Adam optimizer with an initial learning rate 0.002 and MAE as the loss function. For the selection of inner prediction models, we choose a representative STGNN based work D2STGNN~\cite{shao2022decoupled}. We use the default parameters provided in their paper and remove the output layer for subsequent data processing.

\subsection{Experiment results}
We compare the performance of our STMGF and all the baselines of 30 minutes (horizon 6), 60 minutes (horizon 12), 90 minutes (horizon 18), and 120 minutes (horizon 24). The results are demonstrated in \autoref{tab_results}. Our proposed STMGF outperforms the baseline model in all testing metrics. It has advantages especially in long-term prediction. For example, on the PEMS08 dataset, STMGF improves MAE by 8.1\%, RMSE by 6.2\%, MAPE by 3.5\% on Horizon 24. This is related to the temporal multi-granularity method, where multi-granularity timeline reduces the time slices of long-term prediction and improves accuracy. 

\begin{table*}[!ht]
\vspace{-0.2in}
    \centering
    \caption{Prediction results.}
    \label{tab:results}
    \resizebox{\linewidth}{!}{
    \begin{tabular}{l l l l l l l l l l l l l}
    \hline
        Datasets&\multicolumn{12}{c}{\textbf{METR-LA/PEMS08}}\\ \hline
        \multirow{2}{*}{Metrics} & \multicolumn{3}{c}{\textbf{Horizon 6}}  & \multicolumn{3}{c}{\textbf{Horizon 12}}  & \multicolumn{3}{c}{\textbf{Horizon 18}}  & \multicolumn{3}{c}{\textbf{Horizon 24}}\\
        \cmidrule(r){2-4} \cmidrule(r){5-7} \cmidrule(r){8-10} \cmidrule(r){11-13}
         &  MAE& RMSE& MAPE(\%)  &  MAE& RMSE& MAPE(\%)  &  MAE& RMSE& MAPE(\%)  &  MAE& RMSE& MAPE(\%)  \\ \hline\hline
        HA & 8.99/28.38&13.78/42.13&19.99/19.70 & 8.99/28.38&13.78/42.13&19.99/19.70 & 8.99/28.38&13.78/42.13&19.99/19.70 & 8.99/28.38&13.78/42.13&19.99/19.70 \\ 
        ARIMA & 5.07/23.51&11.51/35.92&12.31/14.43 & 6.82/34.14&14.86/26.83&16.92/21.34 & 8.27/45.82&17.13/66.23&20.75/29.28 & 9.46/56.91&18.79/80.43&23.91/37.53 \\ \hline
        DCRNN & 3.08/15.36&6.33/24.22&8.48/10.03 & 3.53/17.14&7.48/26.83&10.28/11.19 & 3.83/18.78&8.20/29.16&11.52/12.38 & 4.04/20.08&8.66/31.09&12.36/13.34 \\ 
        GraphWaveNet & 3.57/16.01&7.24/25.99&10.25/10.71 & 4.39/18.78&8.88/30.33&13.18/12.42 & 4.99/21.98&9.92/35.24&16.26/14.64 & 5.55/25.01&10.62/39.85&18.19/17.46 \\ 
        DSTAGNN & 5.24/15.85&12.91/25.15&10.45/10.07 & 6.72/17.39&15.99/27.64&12.51/11.11 & 7.82/18.74&17.91/29.72&14.16/11.88 & 8.59/20.20&19.04/31.77&15.55/12.48 \\ 
        ASTGCN & 4.80/17.65&8.81/27.82&12.28/12.89 & 5.64/18.59&10.03/29.76&14.45/13.32 & 6.13/19.11&10.64/30.87&15.62/13.58 & 6.52/20.46&11.06/32.53&16.58/14.53 \\ 
        PGCN & 3.17/14.98&6.38/23.84&9.12/9.50 & 3.59/16.20&7.50/26.26&10.81/10.23 & 3.82/17.25&8.00/28.05&11.55/11.02 & 3.99/18.29&8.25/29.60&12.12/12.01 \\ 
        HGCN & 3.38/16.95&6.86/25.69&9.89/11.87 & 3.80/19.04&7.85/28.57&11.61/13.73 & 4.07/21.38&8.37/31.61&12.57/16.22 & 4.29/23.22&8.81/33.89&13.19/18.13 \\ 
        D2STGNN & 2.95/14.49&6.06/24.48&8.15/9.31 & 3.37/15.76&7.15/27.30&9.78/10.12 & 3.63/16.59&7.77/29.00&10.74/10.63 & 3.85/17.32&8.17/30.12&11.48/11.08 \\ 
        PDFormer & 4.15/\textbf{13.80}&9.60/23.58&10.65/9.39 & 4.97/14.88&11.34/25.62&12.89/10.44 & 5.37/15.76&12.16/27.17&13.99/11.40 & 5.83/17.52&12.98/29.03&15.04/12.48 \\ 
        MTGNN & 3.16/15.45&6.45/24.35&8.57/9.93 & 3.60/16.64&7.60/26.43&10.17/10.75 & 3.85/17.87&8.15/28.40&11.22/11.40 & 4.08/19.09&8.58/29.99&12.00/12.66 \\\hline
        STMGF*& 3.05/15.06 & 6.30/24.20 & 8.41/9.81 & 3.43/16.10 & 7.32/26.36 & 10.00/10.40 & 3.67/16.85 & 7.84/27.92 & 11.02/11.18 & 3.87/17.68 & 8.15/29.40 & 11.94/11.71\\
        STMGF   & \textbf{2.94/13.80}&\textbf{6.00/22.96}&\textbf{8.03/9.18} & \textbf{3.33/14.73}&\textbf{7.08/25.23}&\textbf{9.56/9.81} & \textbf{3.56/15.38}&\textbf{7.60/26.43}&\textbf{10.39/10.16} & \textbf{3.77/15.91}&\textbf{7.97/27.23}&\textbf{11.07/10.69} \\ \hline
    \end{tabular}
    }
\vspace{-0.2in}
\end{table*}

\subsection{Ablation study}\label{ablation}
In this subsection, we conduct experiments to verify the effectiveness of important components of our STMGF and change the inner STGNN module to verify the universality of STMGF. We conduct three part ablation experiments of STMGF:

\noindent(1) \textbf{STMGF w/o spatial}. We separately remove the spatial clustering module to ensure that the three layers have the same spatial granularity to verify the effectiveness of the spatial multi-granularity method.

\noindent(2) \textbf{STMGF w/o temporal}.In order to prove the effectiveness of the time clustering method, we make the three layers have the same temporal granularity, which is to use the same length time series for prediction,.

\noindent(3)  \textbf{STMGF w/o historical}. We remove the historical period data matching module and use only recent data for prediction to verify the improvement of prediction results by introducing historical period data.

\begin{figure*}[htbp]
	\centering
	\subfloat[METR-LA.]{\includegraphics[width=.48\linewidth]{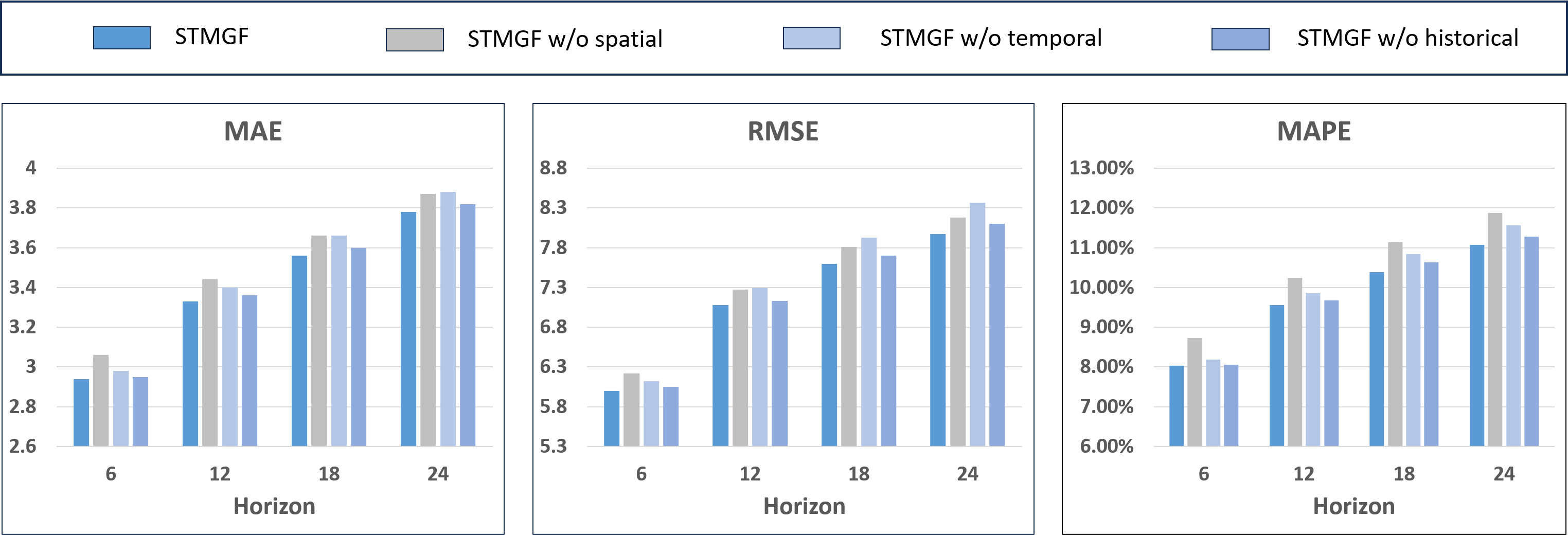}}\hspace{5pt}
	\subfloat[PEMS08.]{\includegraphics[width=.48\linewidth]{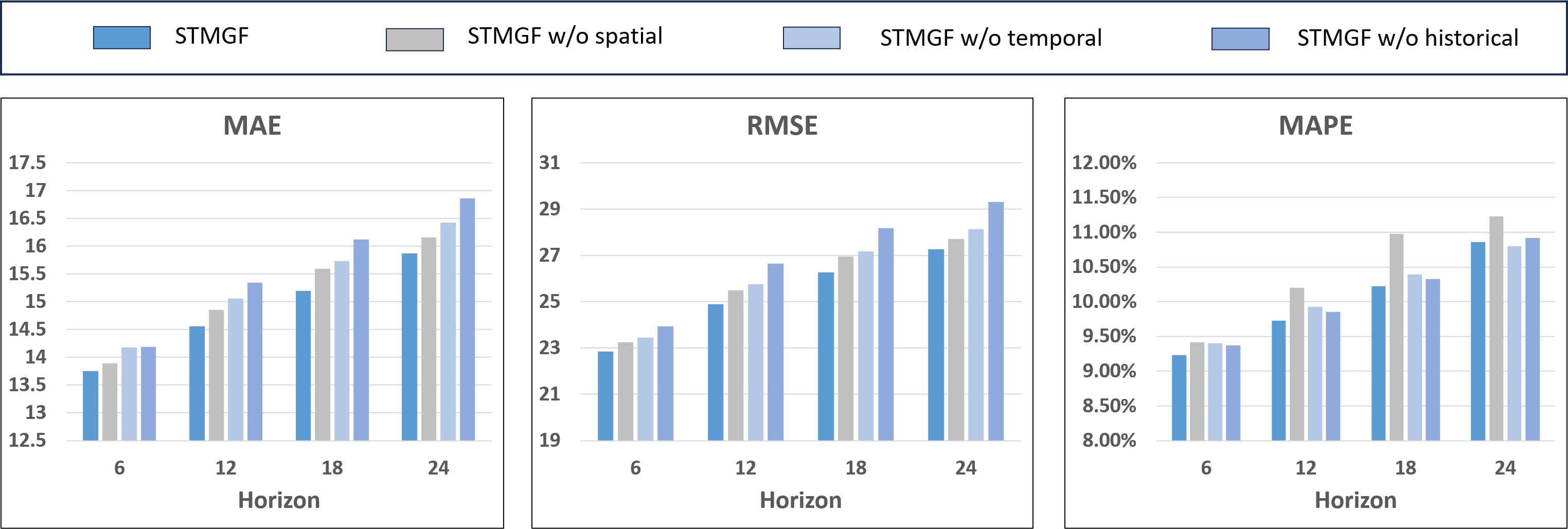}}
    \vspace{-0.1in}
	\caption{Ablation study results.}
    \vspace{-0.3in}
    \label{Ablation}
\end{figure*}

The results are shown in \autoref{Ablation}. The results on two datasets show that STMGF outperforms the ablated model in most cases. This proves the effectiveness of the aforementioned components. To demonstrate that STMGF can enhance the effectiveness of different STGNNs, we further replace the inner STGNN prediction module (D2STGNN\cite{shao2022decoupled}) to MTGNN~\cite{wu2020connecting} (STMGF*). As shown, STMGF* outperforms MTGNN in all cases. STMGF-MTGNN improves 7.3\% MAE, 1.9\% RMSE, and 7.5\% MAPE on the PEMS08 dataset at horizon 24. This further demonstrates the ability of STMGF to improve long-term prediction accuracy and the universality on different inner STGNN models.

\section{CONCLUSIONS}\label{chapter7}
In this paper, we present a novel \textbf{S}patial-\textbf{T}emporal \textbf{M}ulti-\textbf{G}ranularity \textbf{F}ramework for traffic prediction. We apply multi-granularity techniques simultaneously in both the temporal and spatial dimensions, enhancing the ability to capture long-term, long-distance information through hierarchical clustering and attention based interaction mechanisms. We recognize the periodicity in traffic sequences and use the similarity between recent traffic data and historical periodic data to refine the prediction results. Experimental results demonstrate the superiority of our framework in addressing traffic prediction problems and the universality of our framework for various STGNN-based prediction models.

\section*{Acknowledgement}
This work was supported by Shandong Provincial Natural Science Foundation (No ZR2022QF114)

%
%
%
\bibliographystyle{splncs04}
\bibliography{cite}
\end{document}